\documentclass[pdflatex,sn-mathphys-num]{sn-jnl}
\usepackage{multirow}
\usepackage{lineno}
\usepackage{hyperref}
\usepackage[labelsep=space]{caption}
\usepackage{graphicx}%
\usepackage{multirow}%
\usepackage{amsmath,amssymb,amsfonts}%
\usepackage{amsthm}%
\usepackage{mathrsfs}%
\usepackage[title]{appendix}%
\usepackage{xcolor}%
\usepackage{textcomp}%
\usepackage{manyfoot}%
\usepackage{booktabs}%
\usepackage{algorithm}%
\usepackage{algorithmicx}%
\usepackage{algpseudocode}%
\usepackage{listings}%
\usepackage{xcolor}
\usepackage{subcaption}  
\usepackage[capitalise]{cleveref}

\newcommand{\revision}[1]{#1}
\newcommand{\revisisonSecondTour}[1]{#1}

\theoremstyle{thmstyleone}%

\theoremstyle{thmstyletwo}%

\theoremstyle{thmstylethree}%

\begin{document}

\title[Article Title]{Automatic Calibration of a Multi-Camera System with Limited Overlapping Fields of View for \\ 3D Surgical Scene Reconstruction}

\author*[1,2]{\fnm{Tim} \sur{Flückiger}}\email{flueckigertim@gmail.com}

\author[1,2]{\fnm{Jonas} \sur{Hein}}

\author[1]{\fnm{Valery} \sur{Fischer}}

\author[1]{\fnm{Philipp} \sur{Fürnstahl}}

\author[1]{\fnm{Lilian} \sur{Calvet}}

\affil[1]{\orgdiv{Research in Orthopedic Computer Science}, \orgname{University Hospital Balgrist, University of Zurich}, \orgaddress{\country{Switzerland}}}
\affil[2]{\orgdiv{Computer Vision and Geometry}, \orgname{ETH Zurich}, \orgaddress{\country{Switzerland}}}

\abstract{\textbf{Purpose:}
The purpose of this study is to develop an automated and accurate external camera calibration method for multi-camera systems used in 3D surgical scene reconstruction (3D-SSR)\revision{, eliminating the need for operator intervention or specialized expertise. The method specifically addresses the problem of limited overlapping fields of view caused by significant variations in optical zoom levels and camera locations.} \\
\textbf{Methods:}
We contribute a novel, fast, and fully automatic calibration method based on the projection of multi-scale markers (MSMs) using a ceiling-mounted projector. MSMs consist of 2D patterns projected at varying scales, ensuring accurate extraction of well distributed point correspondences across significantly different viewpoints and zoom levels. Validation is performed using both synthetic and real data captured in a mock-up OR, with comparisons to traditional manual marker-based methods as well as markerless calibration methods. \\
\textbf{Results:}
\revision{The method achieves accuracy comparable to manual, operator-dependent calibration methods while exhibiting higher robustness under conditions of significant differences in zoom levels. }
\revision{Additionally, we show that state-of-the-art Structure-from-Motion (SfM) pipelines are ineffective in 3D-SSR settings, even when additional texture is projected onto the OR floor.}\\
\textbf{Conclusion:} The use of a ceiling-mounted entry-level projector proves to be an effective alternative to operator-dependent, traditional marker-based methods, paving the way for fully automated 3D-SSR.
}


\maketitle

\section{Introduction}\label{Introduction}

3D surgical scene reconstruction (3D-SSR) involves computing a 3D representation of all relevant entities during surgery, including the anatomy shape and texture, surgical instrument poses, and body and hand poses of the medical staff. It can facilitate error prevention, operative performance assessment, formative feedback by avoiding the need for manual review and assessment of surgical video \citep{MascagniASAMWAR22,RazaVDK22}.
For tasks such as automatic workflow analysis and activity recognition, the use of geometric representations have recently been shown to outperform image-based representations while requiring less labeled data \citep{pmlr-v227-hamoud24a}.
In surgical navigation and robotic surgery, 3D-SSR can help to reduce the sim-to-real gap by providing accurate and realistic 3D reconstructions of surgical scenes in which robots and machine learning (ML)-based applications can be trained before being deployed in the real world \citep{Gumbs21,Barnoyetal21,GumbsGBCSFIHPE22,JecklinJFFE22}.
Optical cameras remain the preferred solution for capturing 3D information related to medical staff, including their pose, surgeon hand movements, surgical instrument positions, and anatomy shape and texture. They are widely used for their ability to capture detailed visual information in a non-invasive manner while recording events in real-time. We focus on 3D-SSR based on a multi-camera system in this work.

3D-SSR requires the fusion of information from the cameras into a common spatio-temporal representation that accurately describes the state of the surgery.
A prerequisite for this fusion is the external calibration of the cameras, which involves computing their spatial relationships, namely their positions and orientations in 3D space, referred to as camera poses. External camera calibration has been a long-standing problem in the computer vision research community that is often considered solved, especially in acquisition setups allowing for dense viewpoint distribution. However, it remains highly challenging in various environments, such as an operating room (OR). In the 3D-SSR acquisition scenario, three main challenges arise.

\begin{figure}[t]
    \centering
    \begin{minipage}{0.33\textwidth}
        \centering
        \includegraphics[width=\textwidth]{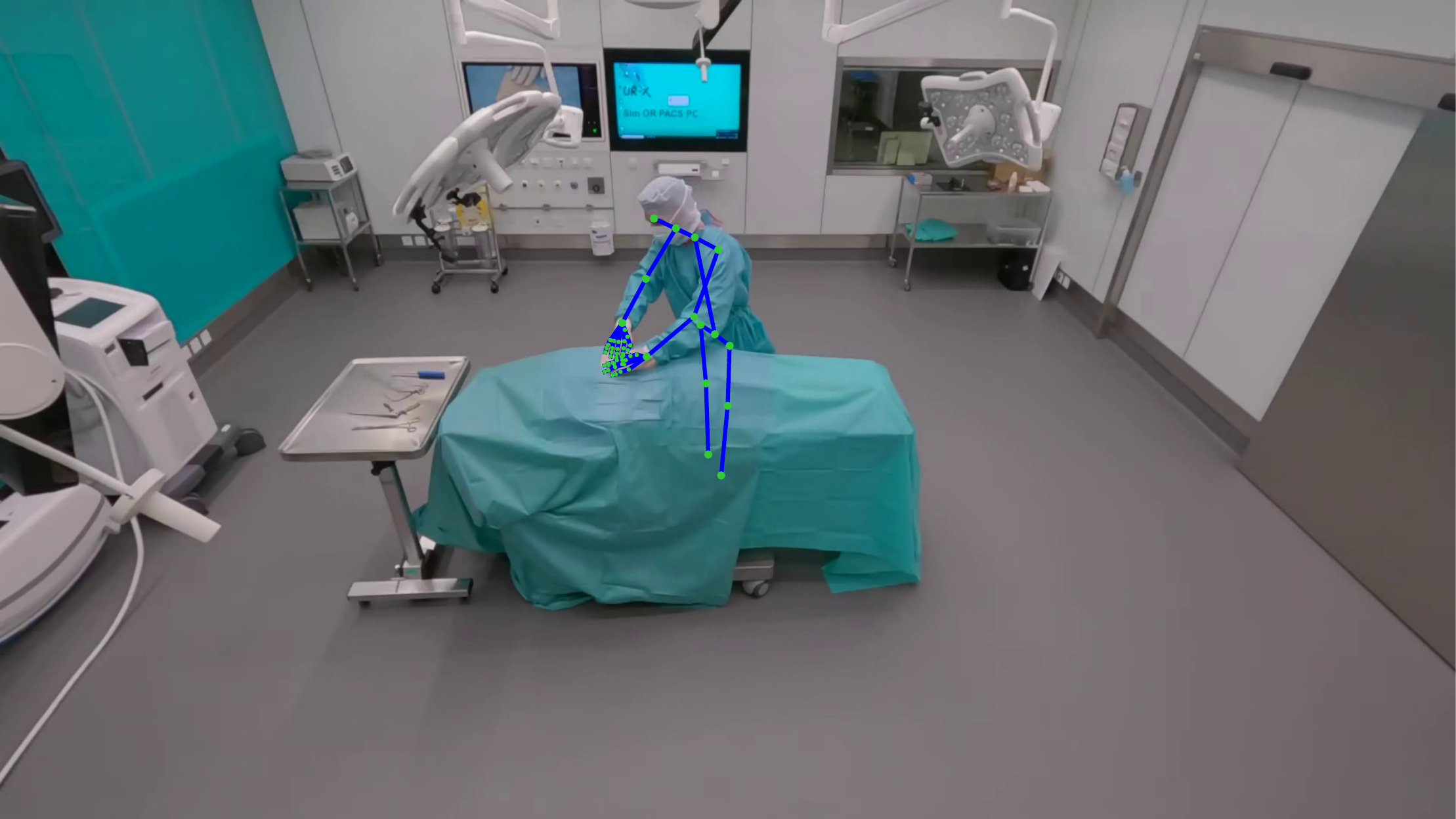}
    \end{minipage}%
    \begin{minipage}{0.33\textwidth}
        \centering
        \includegraphics[width=\textwidth]{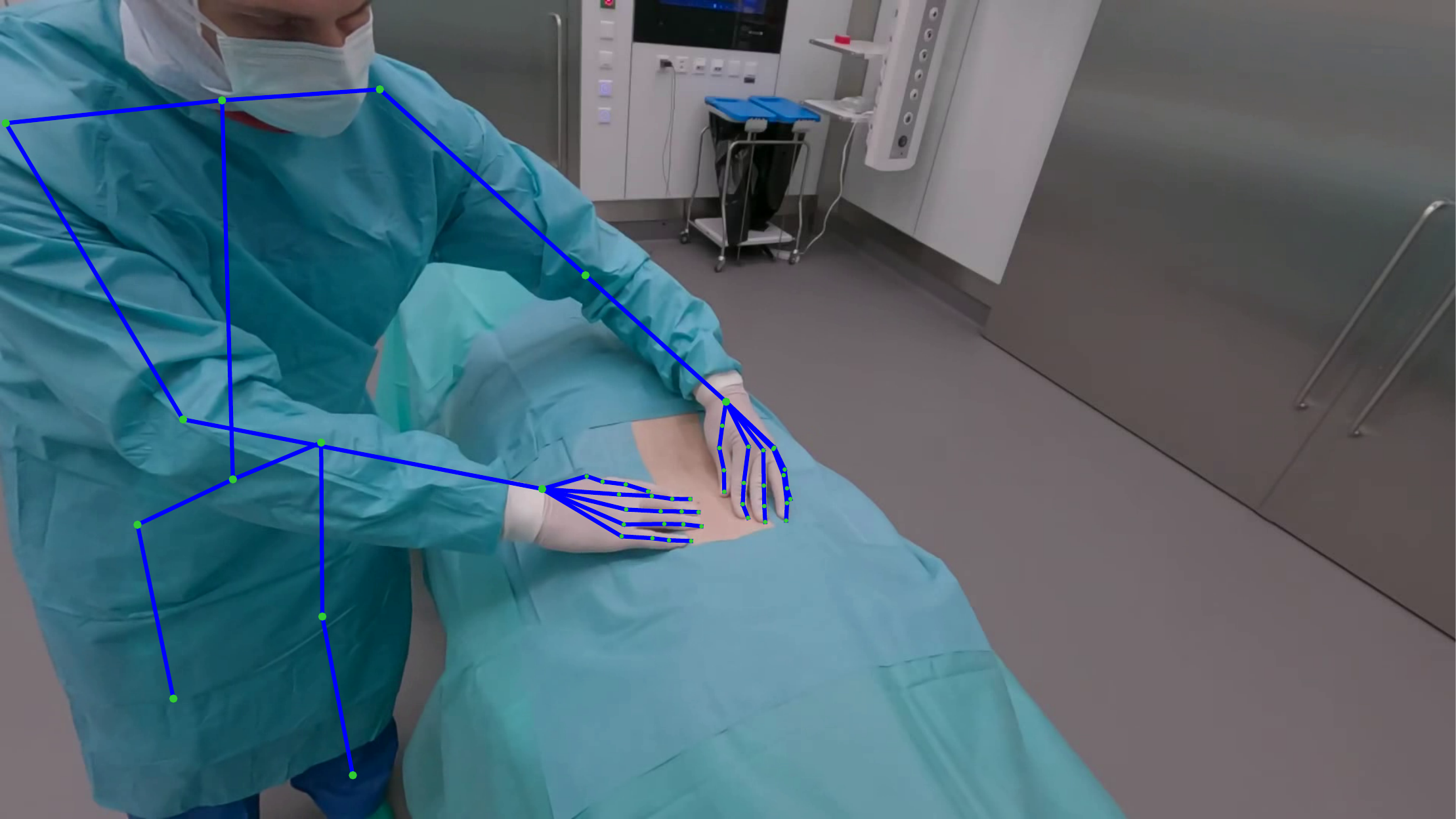}
    \end{minipage}%
    \begin{minipage}{0.33\textwidth}
        \centering
        \includegraphics[width=\textwidth]{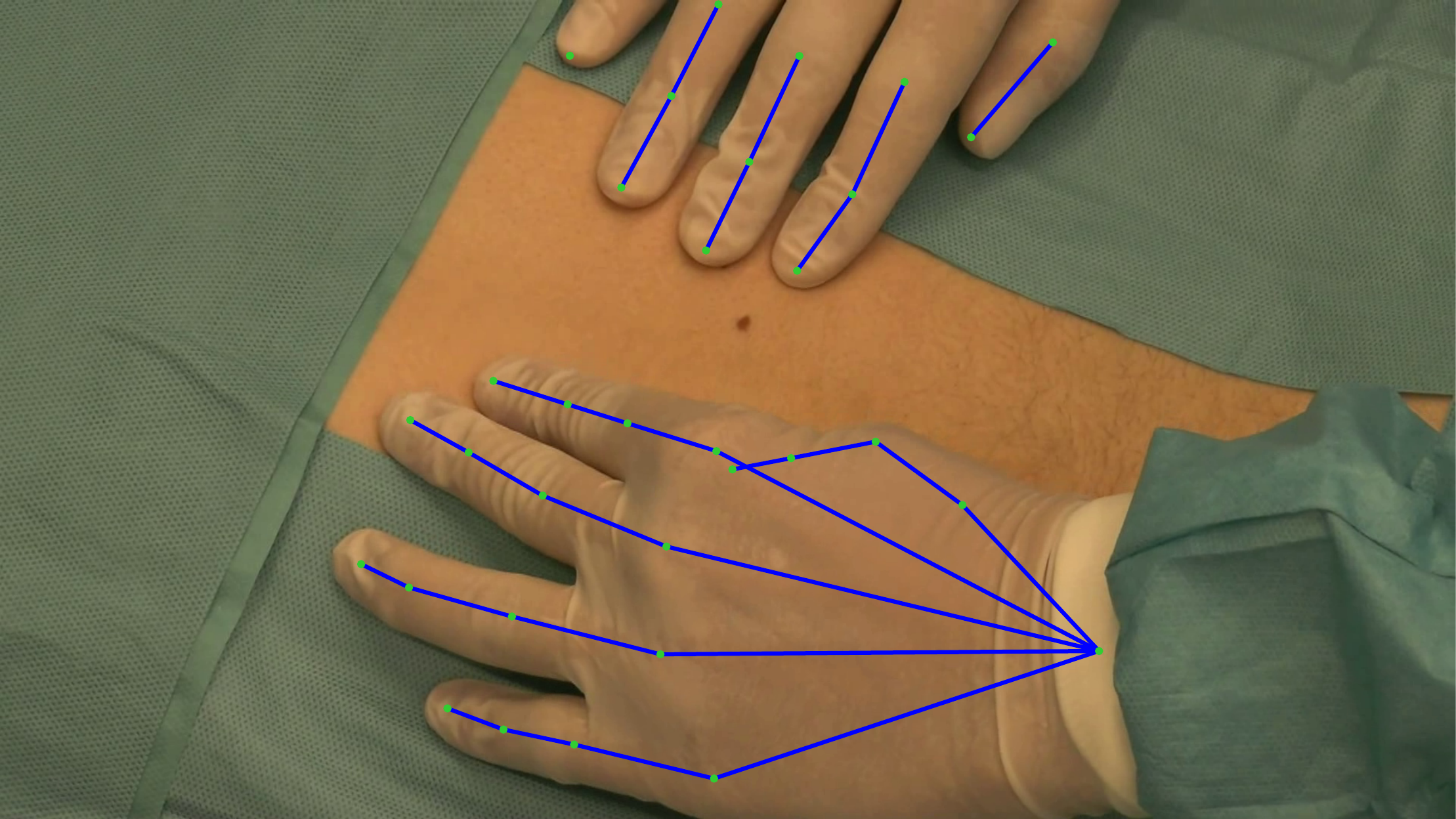}
    \end{minipage}%

    \begin{minipage}{0.33\textwidth}
        \centering
        \includegraphics[width=\textwidth, trim=0 50 0 50, clip]{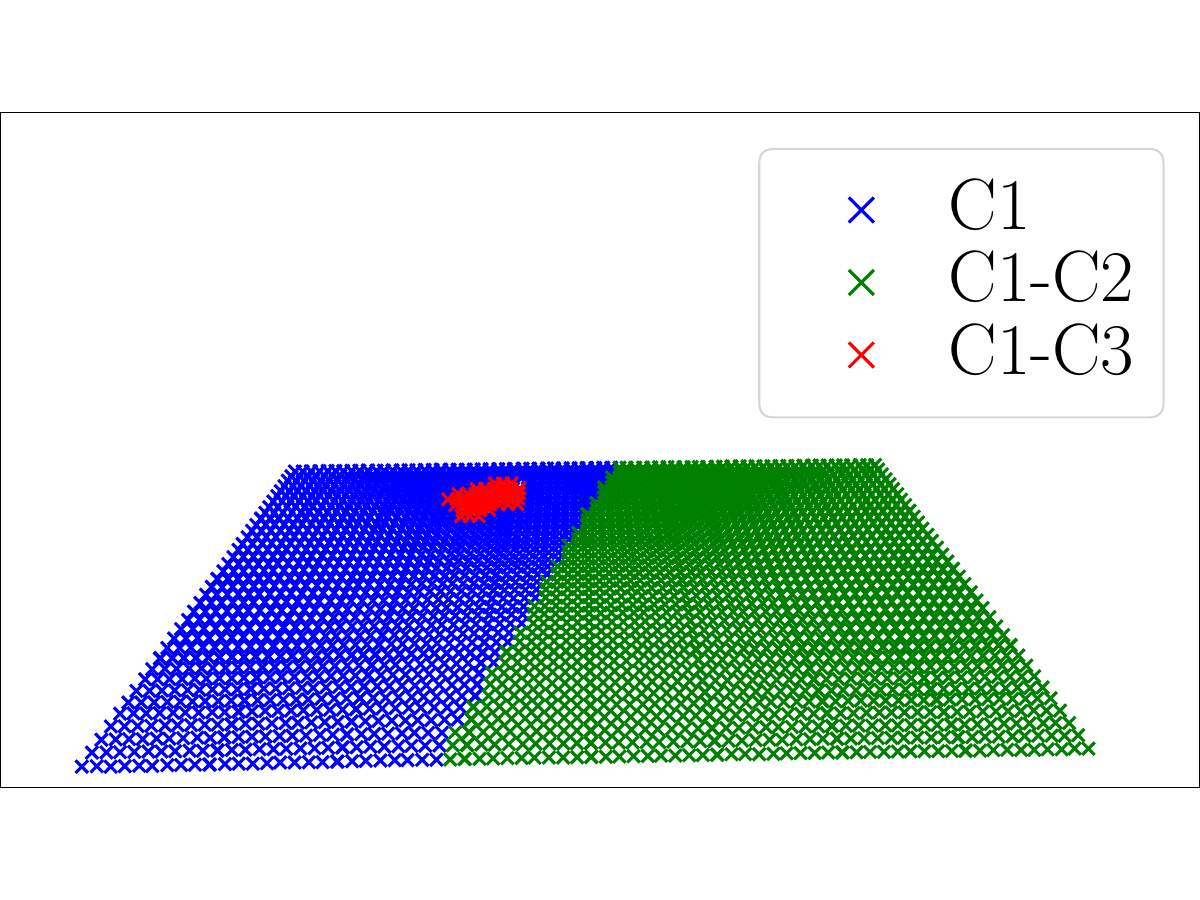}
    \end{minipage}%
    \begin{minipage}{0.33\textwidth}
        \centering
        \includegraphics[width=\textwidth, trim=0 50 0 50, clip]{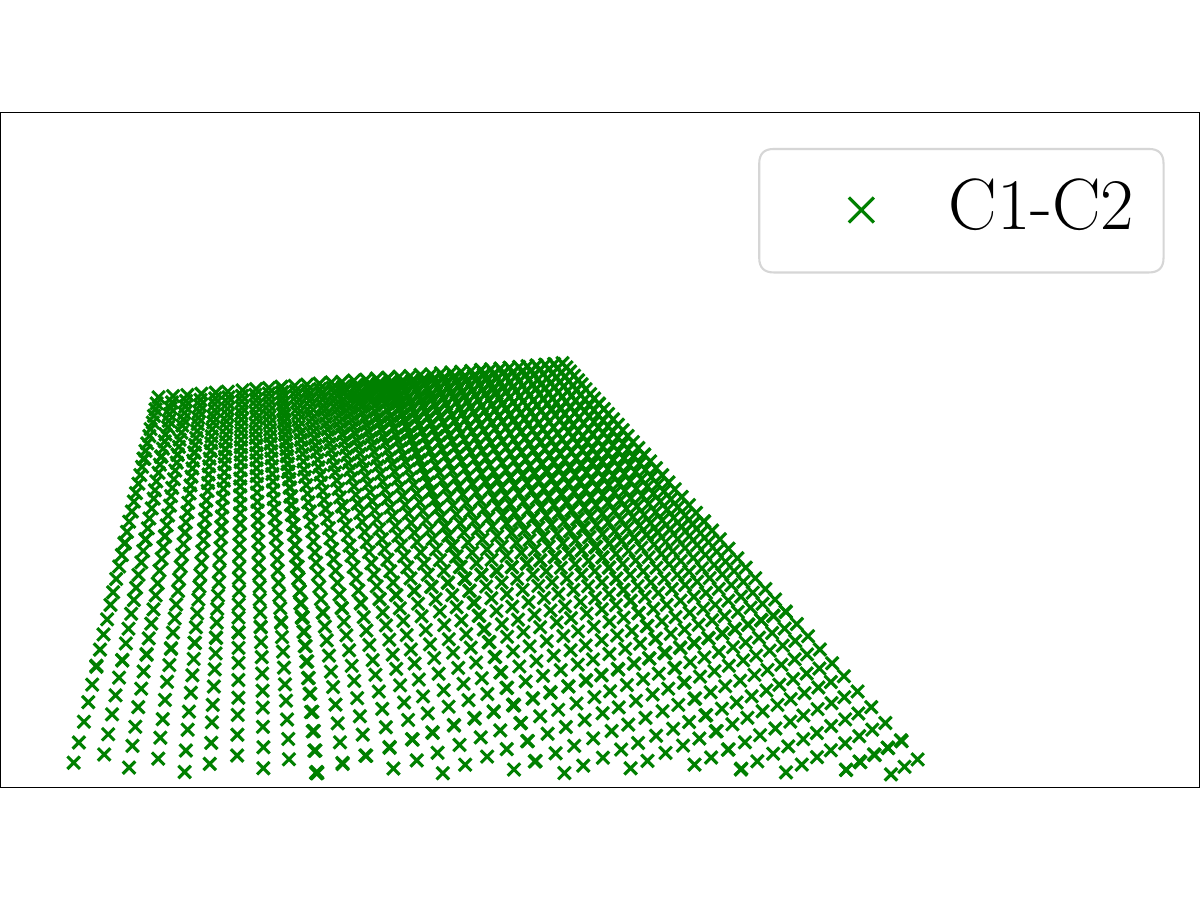}
    \end{minipage}%
    \begin{minipage}{0.33\textwidth}
        \centering
        \includegraphics[width=\textwidth, trim=0 50 0 50, clip]{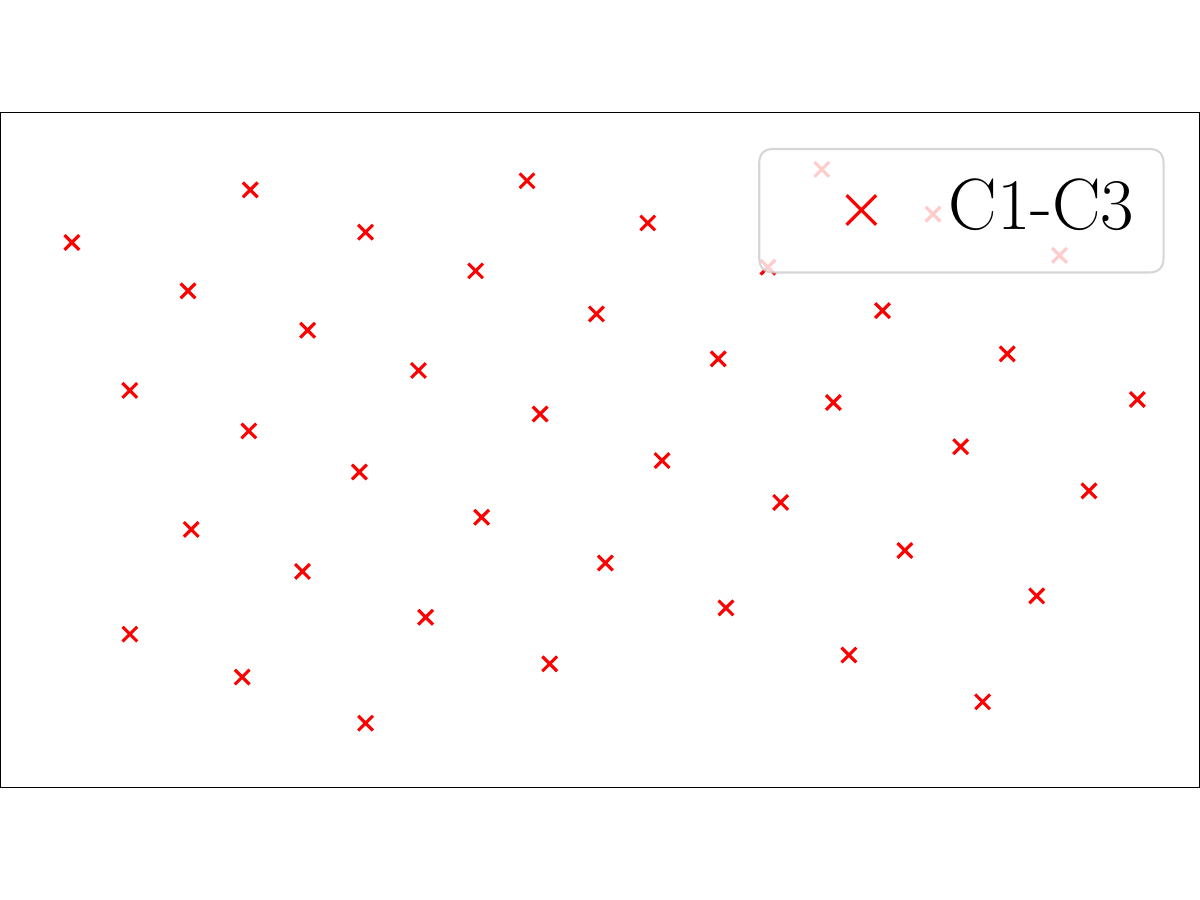}
    \end{minipage}%

    \vspace{0.1cm}

    \caption{Top: Examples of camera viewpoints used for 3D surgical scene reconstruction. From left to right: far-field camera C1 (ceiling-mounted GoPro), near-field camera C2 (surgical lamp-mounted GoPro), and near-field camera C3 (ceiling-mounted Canon CR-N300), zoomed in on the surgical field.
\mbox{Bottom}: Inter-view point correspondences provided by our multi-camera calibration system, observed from the same viewpoints.
From left to right: all correspondences seen by C1 (blue, green, and red), correspondences between C1 and C2 (green), and correspondences between C1 and C3 (red).
The surgeon's body and hand poses, represented by blue lines in the first row, are successfully reconstructed using the estimated camera poses}
    \label{fig:intro}
\end{figure}

The first challenge is sparsely sampled viewpoints (also referred to as \textit{sparse-view} or \textit{wide-baseline}) which leads to failures of Structure-from-Motion (SfM) pipelines \citep{schonberger2016structure, pan2024glomap} due to the lack of reliable point correspondences across views.
The second challenge involves complex materials (glossy, metallic, and mirrored surfaces) and poorly textured surfaces in the OR which also defeat standard SfM pipelines.
The use of easily identifiable markers with known geometry, placed on a manufactured planar calibration pattern (e.g., standard checkerboard, ChArUco \citep{garrido2014automatic}, AprilTag \citep{apriltags}, ArUco \citep{aruco}), is often employed to address these technical limitations. Typically, a calibration pattern is moved by an operator within the volume intersection of the camera frustums, providing a large set of inter-view point correspondences.
This family of methods is now established as the gold standard for calibrating multi-camera systems \citep{Bouguet2010, Rehder2016kalibr}.
\revision{Other methods, such as those utilizing manually moved LEDs or laser pointers \cite{led1, led2} as calibration objects, are also employed to obtain inter-view point correspondences.}
\revision{However, all these methods are time-consuming, operator-dependent and require a certain level of expertise.
Such limitations make them impractical in busy environments like operating rooms, where time between surgeries is limited, and hospital staff very rarely, if ever, have the expertise required for computer vision or calibration techniques.
} 
The third challenge in calibrating a multi-camera system for 3D-SSR lies in \revision{the weak overlap between fields of view, particularly due to significant changes in camera location and zoom level relative to the size of the OR. For example, the small red region in \cref{fig:intro}, which represents the field of view of camera C3 (associated to the top right image) as seen by camera C1, corresponding to an appearing scale change of 17x, is a typical example of this issue.
This problem arises from the diverse nature of objects to be reconstructed—ranging from wounds typically 5 to 30 cm in length to individuals moving within the OR, which can encompass up to 200 cubic meters—imposing high variations in camera positions and zoom levels. 
By \textit{limited overlapping fields of view}, we refer in the sequel to limited overlaps arising from significant changes in image scale—caused by substantial variations in position, zoom level, or a combination of both. This differs from limited overlap due to diverging camera orientations, which is not inherently related to 3D-SSR and has been specifically addressed in the literature \citep{simpleCalibNonOverlap, board3}}.

\revision{We propose a novel method that is fully automated, fast, accurate, and designed to address the challenges posed by 3D-SSR.}
The key idea of the method is the use of a ceiling-mounted projector that projects arrays of multi-scale markers (MSMs) onto the floor. An MSM consists of 2D patterns projected at varying scales, with centers that remain invariant to scale changes. This ensures that all projected markers are successfully extracted and accurately localized from all viewpoints, even when handling extreme variations in fields of view (see left and right in \cref{fig:intro}).
Multiple arrays of MSMs are projected, providing a large number of uniformly distributed point correspondences across all views. The process is fully automated, operates with a standard entry-level projector, does not require projector calibration, and exploits static marker projections, ensuring highly accurate extraction of correspondences without facing motion blur or temporal synchronization issues.
The method is fast, with a required projection duration of about 70 seconds, allowing for last-minute adjustments to camera positions, particularly those mounted on movable lamps. Our multi-camera calibration method accurately calibrates all cameras in our experiment with average reprojection errors of 0.28 pixel, while nicely dealing with large scale variations imposed by surgical settings. 
In contrast, off-the-shelf ChArUco \citep{garrido2014automatic}, checkerboard-based calibration method \citep{Bouguet2010, Rehder2016kalibr}, and state-of-the-art SfM pipelines \citep{schonberger2016structure, pan2024glomap} fail to recover all camera poses.
The source code is publicly available at \url{https://github.com/tflueckiger/calib-proj}.

\begin{figure}[t]
    \centering
   
    \begin{minipage}{0.429\textwidth}
        \centering
        \includegraphics[width=0.98\textwidth]{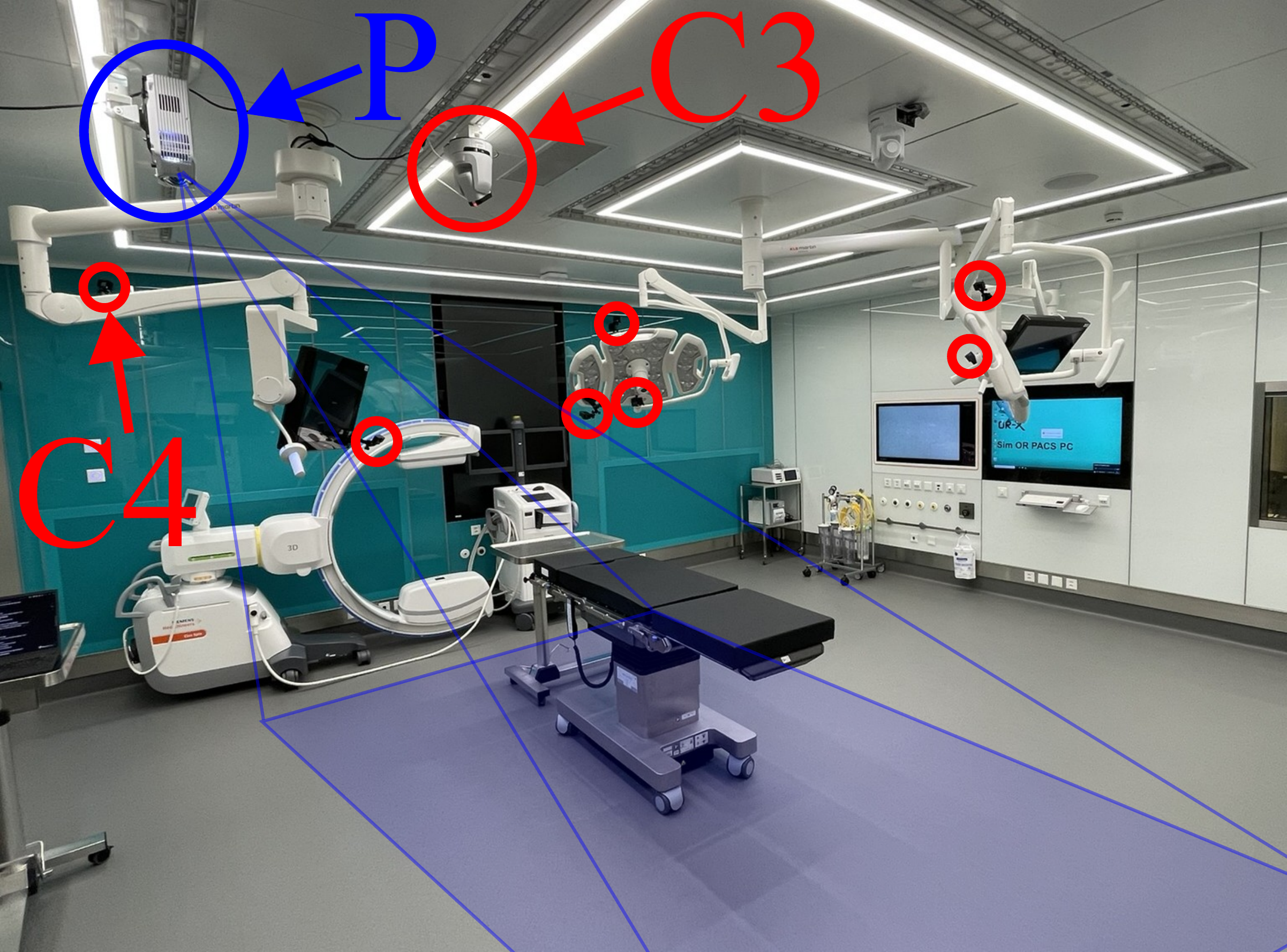}
        \subcaption{}
    \end{minipage}%
 \begin{minipage}{0.57\textwidth}
  
    \begin{minipage}{0.49\textwidth}
        \centering
        \includegraphics[width=0.98\textwidth]{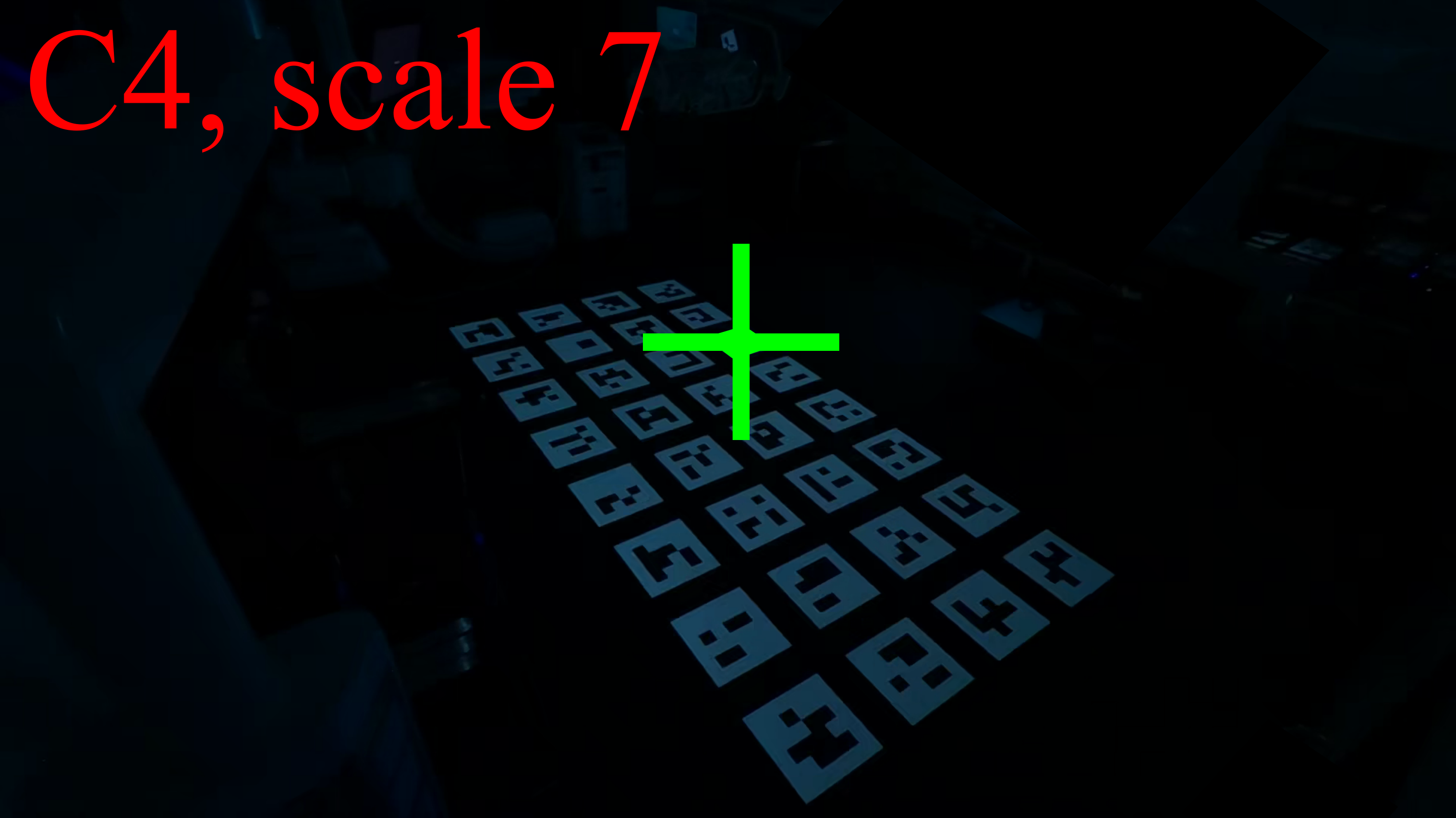}
    \end{minipage}%
    \begin{minipage}{0.49\textwidth}
        \centering
        \includegraphics[width=0.98\textwidth]{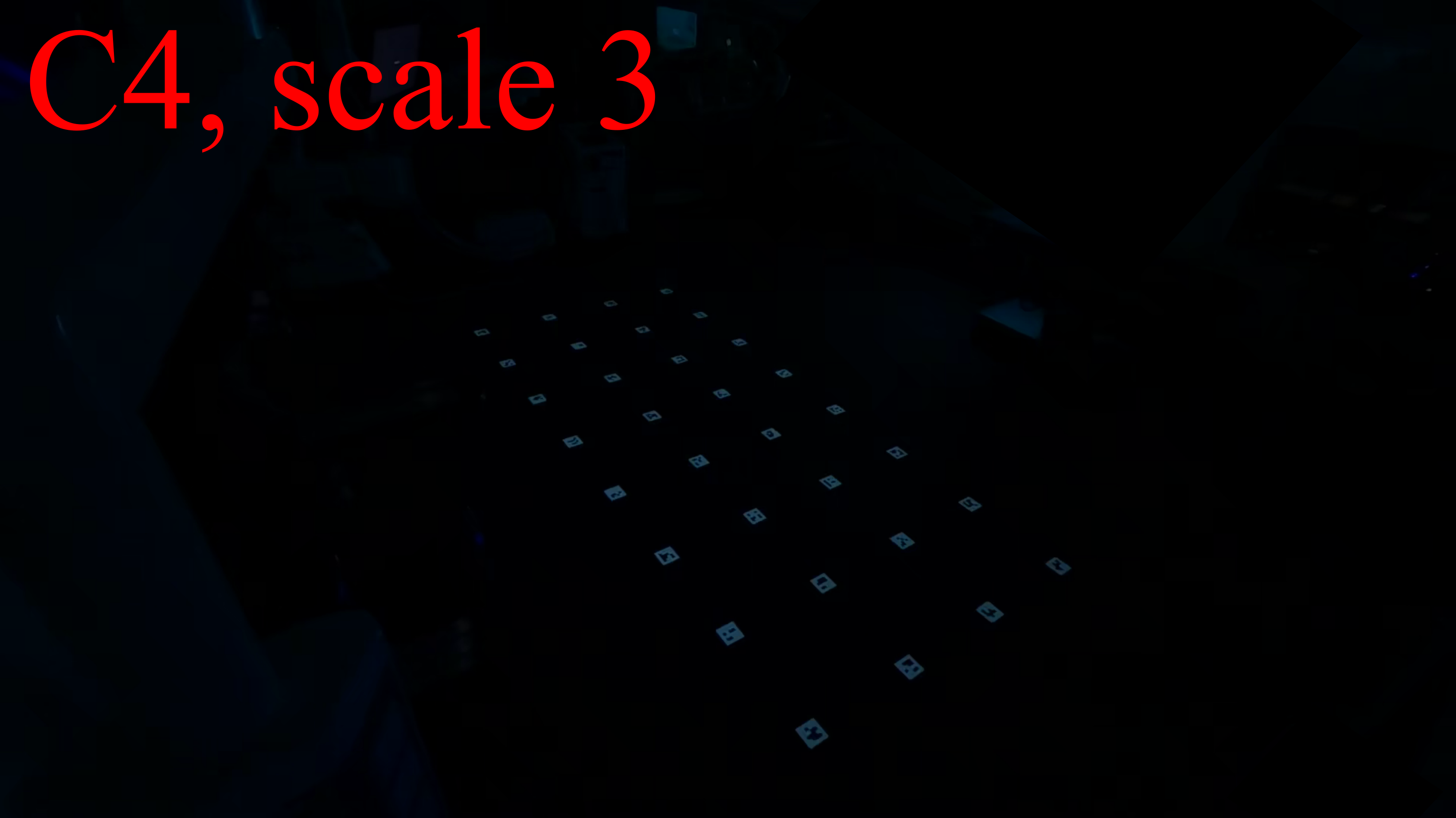}
    \end{minipage}%

    \begin{minipage}{0.49\textwidth}
        \centering
        \includegraphics[width=0.98\textwidth]{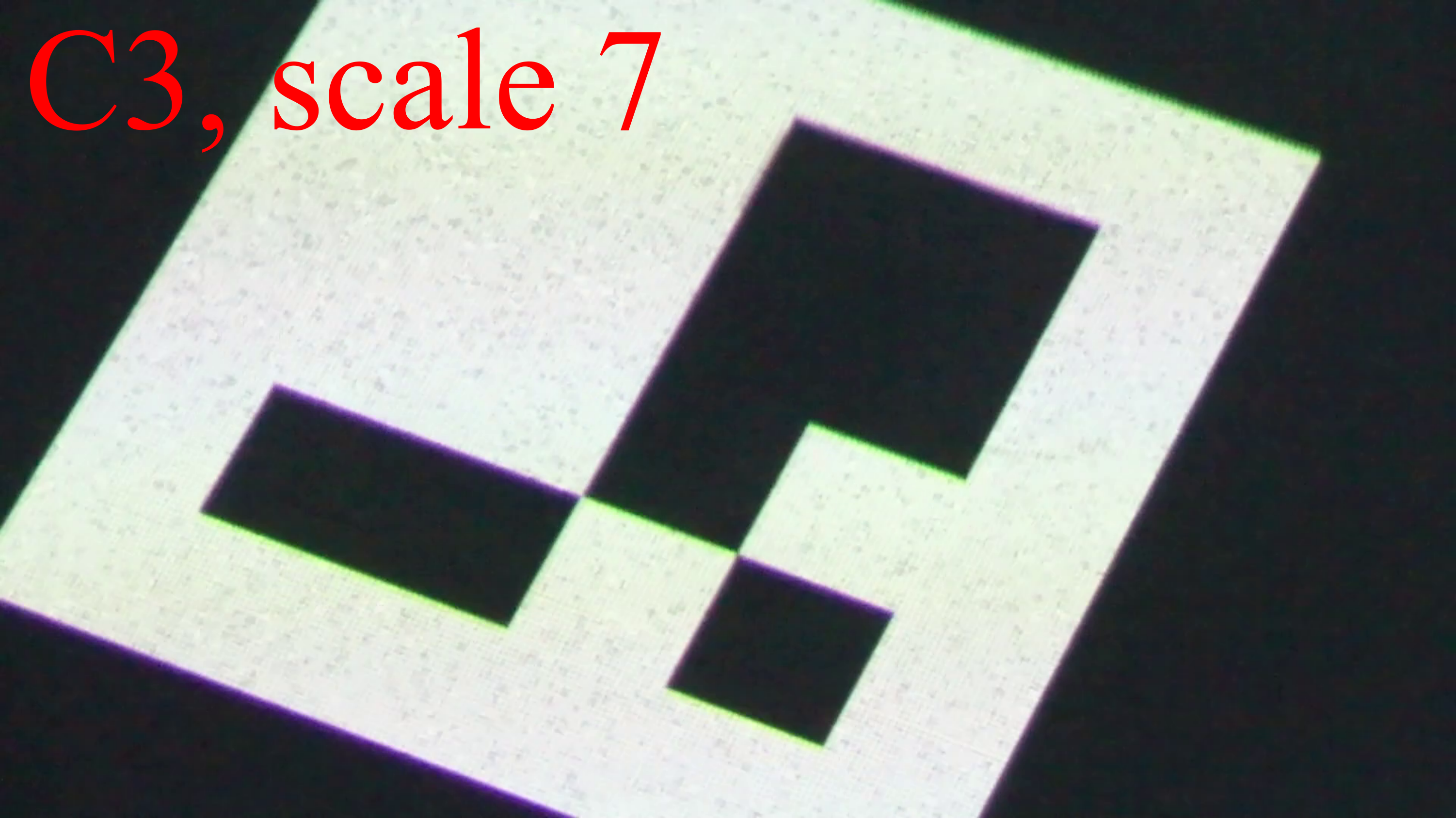}
    \end{minipage}%
    \begin{minipage}{0.49\textwidth}
        \centering
        \includegraphics[width=0.98\textwidth]{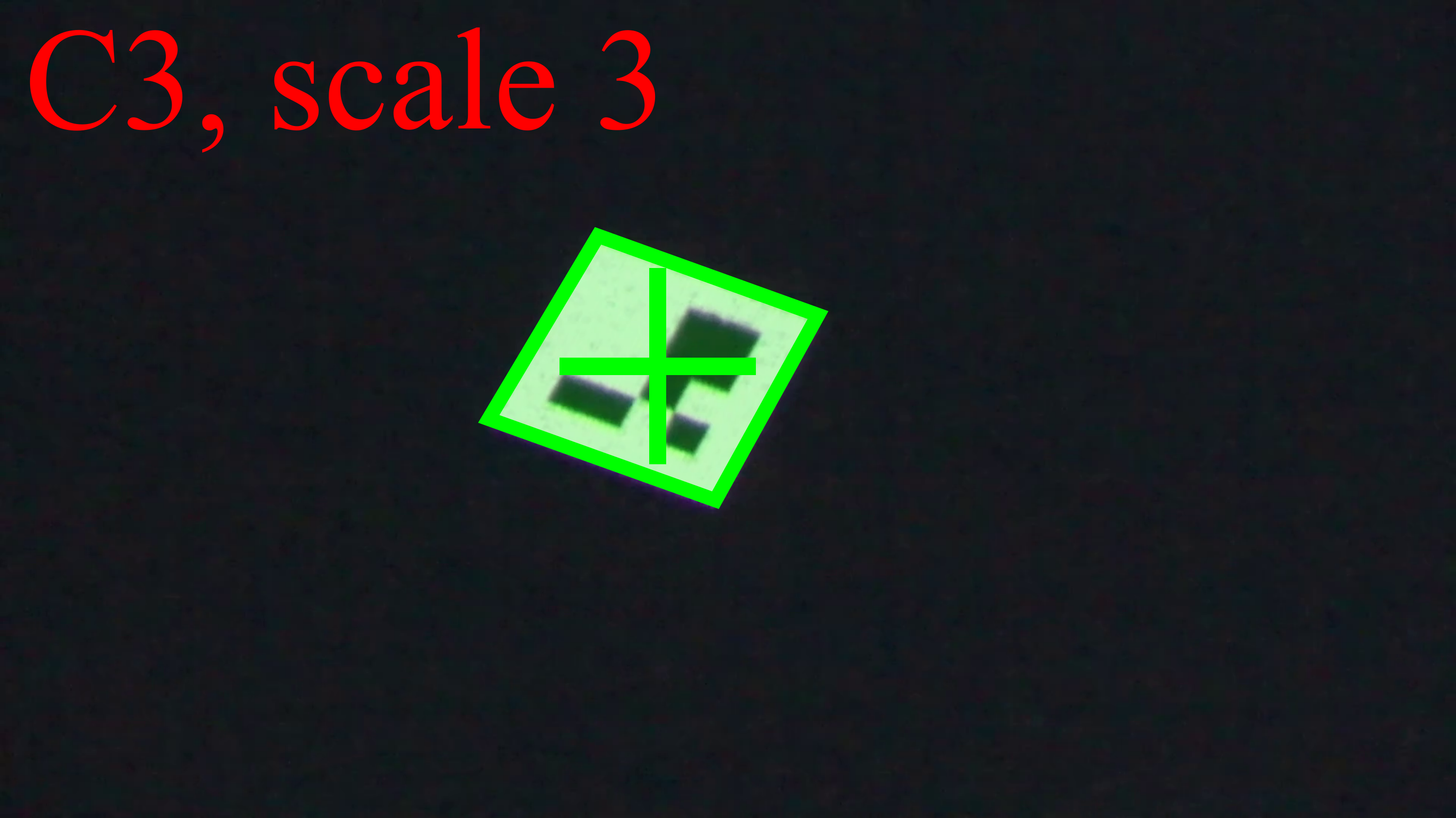}
    \end{minipage}%
    \subcaption{}
\end{minipage}

    \caption{(a) Overview of the proposed setup. Our multi-camera calibration system employs an entry-level projector (P, projection in blue). Cameras are highlighted in red. 
In this example, GoPro cameras placed in far field and in near field on surgical lamps, and a zoomed in Canon CR-N300 (C3) are employed.
(b) Images from a far-field camera (C4, top) and a near field camera (C3, bottom) during the projection of the proposed MSM.
Top: MSM is successfully detected (green cross) in the far-field view for a large scale projection (left) and not detected otherwise (right).
Bottom: MSM is not detected in the near-field view for a large scale projection (left) and successfully detected otherwise (green cross in the right image), therefore providing one point correspondence between C3 and C4 through their shared imaged center. This operation is repeated, providing a set of dense and uniformly distributed point correspondences
}
    \label{fig:methodo_fig}
\end{figure}
The rest of this work is organized as follows. The proposed calibration method is described in \cref{sec:Methods} in which the notion of MSM is introduced. \cref{sec:experiments} extensively evaluates the method and compares it with gold standard marker-based methods and state-of-the-art SfM pipelines.
Finally, conclusions are presented in \cref{sec:Conclusion}.

\section{Methods}\label{sec:Methods}

\subsection{Multi-scale markers}
\label{subsec:MSM}

One of the key challenges posed by 3D-SSR is obtaining correspondences across views under highly varying viewpoints and zoom levels (see \cref{fig:intro}).
To address this challenge, we propose replacing traditional planar markers, which are placed on a calibration pattern moved by an operator, with a series of planar patterns projected onto the OR floor using a ceiling-mounted projector (see \cref{fig:methodo_fig}).

In the following, a \textit{marker} is defined as a set of pattern projections produced by a projector onto the floor, which is assumed to be piecewise planar.
These projections share a common center, referred to as the \textit{marker center}.
Unlike manufactured markers with fixed shapes, typically detected from a single image, the proposed marker is detected from a video sequence. Its varying appearance, offered by the sequence of projections, is designed to optimize detection rates under challenging camera viewpoints and zoom levels.

Concretely, the collection $\mathcal{Q}$ of patterns being projected is defined as images of a Euclidean pattern $\mathcal{P}$
through a set of planar homographies $\mathcal{H}_{\mathbf{p}}$
verifying that the center $\mathbf{p}$ of $\mathcal{P}$ is left unchanged, written as:
\begin{equation}
\mathcal{Q} := \{ h_{\mathbf{p}}(\mathcal{P}) \mid h_{\mathbf{p}} \in \mathcal{H}_{\mathbf{p}} \},
\end{equation}
where $\mathcal{H}_{\mathbf{p}} := \{h_{\mathbf{p}} \in \text{GL}(3, \mathbb{R}) \mid {\mathbf{p}} = h_{\mathbf{p}}({\mathbf{p}}) \}$.
The set of homographies $\mathcal{H}_{\mathbf{p}}$ can, for example, be chosen such that a subset of them compensates for the perspective distortions caused by the projection onto the floor, combined with the projection onto the cameras' image plane. This property is particularly desirable to optimize detection rates, for example in cases of grazing viewpoints.
If the set of homographies $\mathcal{H}_{\mathbf{p}}$ is sufficiently large and its homographies are well-sampled, it is possible to ensure that, for each viewpoint, at least one imaged projection is detected, thereby providing correspondences of their imaged centers even under extreme viewpoint changes.

Without loss of generality, we focus in this work on homography samplings 
in $\mathcal{S}_{\mathbf{p}} \subset \mathcal{H}_{\mathbf{p}}$ which represents the set of planar scaling transformations centered at ${\mathbf{p}}$, written as:
\begin{equation}
\mathcal{S}_{\mathbf{p}} := \left\{ s : \mathbb{R}^2 \to \mathbb{R}^2 \mid s(\mathbf{x}) = \lambda (\mathbf{x} - \mathbf{p}) + \mathbf{p}, \, \lambda \in \Lambda \subset \mathbb{R}_+ \right\}
\end{equation}
with $\Lambda$ being the set of selected scale factors.
Our motivation is twofold. First, it primarily addresses what we consider as the most critical challenge in 3D-SSR: large variations in scale (see \cref{fig:intro}).
Second, the proposed sampling $\Lambda$ is unidimensional, resulting in a small number $|\mathcal{S}_{\mathbf{p}}|$ of pattern projections, which leads to a short projection duration.
The proposed multi-scale marker is thus defined as the collection of patterns $\mathcal{Q}$ projected onto the floor, written as: 
\begin{equation}
   \text{MSM} := \{  g_{\text{proj}}(h_{\mathbf{p}}(\mathcal{P})) \mid h_{\mathbf{p}} \in \mathcal{S}_{\mathbf{p}} \} 
\end{equation}
where $g_{\text{proj}}$ is the homography from the projected pattern $h_{\mathbf{p}}(\mathcal{P})$ to the floor plane, the center of which is denoted by $\mathbf{X} \in \mathbb{R}^3$ in the following.

\noindent \textbf{Euclidean pattern selection}
We recall that our objective is to establish point correspondences between views. This places specific requirements on the choice of the Euclidean pattern $\mathcal{P} \in \left\{ \phi : \mathbb{R}^2 \to [0, 1] \right\}$, which must meet two key criteria.
The first criterion is that $\mathcal{P}$ allows for the computation of a point $\mathbf{c}$, referred to as its \textit{center}, such that $\mathbf{c} = f(\mathcal{P})$,
where $f$ denotes a detection function.
The second criterion is that when $f$ is applied to an image of $\mathcal{P}$ transformed by any homography $g$, it returns the imaged center, namely $f(g(\mathcal{P})) = g(f(\mathcal{P}))=g(\mathbf{c})$. 
By choosing the marker center $\mathbf{p}$ such that $\mathbf{p}=\mathbf{c}$, this second criterion ensures the detection of any image of the 
marker center, since $f(g_{cam}(g_{proj}(h_{\mathbf{p}}(\mathcal{P})))=g(\mathbf{p})$, where  
$g_{\text{cam}}$ is the homography from the floor plane to the image plane of the camera.
Among the patterns that satisfy these two criteria, a square is one example, where the imaged center can be computed using the intersection of its imaged diagonals, which is a projective invariant, or two (or more) concentric circles \citep{concentriccircles}.

\subsection{Calibration algorithm}
\label{subsec:alg}
We now present the algorithm used for solving the external calibration of a multi-camera system using 2D-2D correspondences delivered by the detected imaged MSMs. The problem is that of finding a set of camera poses that best explain the extracted imaged MSMs centers. It is stated as a non-linear least squares optimization problem, commonly known as bundle adjustment (BA), whose solution is

\begin{equation}
\label{eq:BA_free}
    \arg\min_{\{\mathbf{P}_c\}, \{\mathbf{X}_k\}} \sum_{c} \sum_{k} v_{c,k} \left\|  \pi \left( \mathbf{K}_c, \mathbf{P}_c, \mathbf{X}_k  \right) - \mathbf{x}_{c,k}  \right\|_2^2
\end{equation}
where $\mathbf{P}_c$ are the camera poses parameters (parametrizing the rotation and translation), $\mathbf{K}_c$ the (known) camera intrinsics, $\mathbf{X}_k$ the center of the MSM $k$, $\mathbf{x}_{c,k}$ its image in the camera $c$, $v_{c,k}$ equal to 1 if $\mathbf{X}_k$ is observed in camera $c$, 0 otherwise, and $\pi$ the projection function of the camera model. 

We solve \cref{eq:BA_free} in an incremental manner similarly to standard SfM pipelines \citep{schonberger2016structure}. The main difference lies in the initialization of the reconstruction, which consists of computing the relative camera motion of a well-chosen initial camera pair. 

Indeed, in our camera calibration scenario, correspondences are images of coplanar points which represent a degenerate case for estimating relative camera motion using standard epipolar geometry, specifically via the fundamental or essential matrix \citep{Hartley2003MVG}. Instead, we compute the camera motion of the initial camera pair by decomposing the inter-image homography \citep{Hartley2003MVG}, which is the rigorous approach for handling coplanar or nearly coplanar points. The homography is estimated through RANSAC (using a threshold of 3 pixels) on normalized correspondences. 

The initial camera pair is selected based on the view score borrowed from the SfM pipeline \citep{schonberger2016structure}. This score represents the number and distribution of correspondences in a camera. We choose the initial camera pair as the one maximizing the score of both cameras among all possible camera pairs.

Once the initialization is complete, the incremental reconstruction proceeds as follows. The camera with the highest view score, based on correspondences with already triangulated points, is selected. Its pose is then estimated using the PnP algorithm, new points are triangulated, and intermediate bundle adjustment is applied to the set of already reconstructed cameras.

\section{Experiments}
\label{sec:experiments}

\subsection{Calibration on synthetic data}
\label{subsec:xp_calib_synth}

We conducted a preliminary experiment to verify that the spatial distribution of 3D points located on the floor does not negatively impact calibration quality compared to the \textit{gold standard checkerboard} calibration method \citep{Zhang2000CameraCalibration, Bouguet2010, Rehder2016kalibr}, which uses 3D points within the volume intersection of the camera frustums, obtained using a moving calibration board.

\noindent \textbf{Cameras} Our synthetic setup consists of 6 far-field cameras and 4 near-field cameras. 
The far-field and near-field cameras are positioned on two horizontal circles. 
The far-field circle has a radius of 2.8 meters and a height of 2.8 meters, while the near-field circle has a radius of 1.2 meters and a height of 1.4 meters. 
In each simulation, the cameras are positioned randomly on their respective circle while constraining the optical axis to point at the center of the scene.

\noindent \textbf{Point distributions}
We compare the calibration accuracy of the estimated camera poses across three scenarios, each associated with a different 3D point distribution. In each scenario, we sample 3D points and generate synthetic 2D observations by projecting them onto the camera image planes, adding Gaussian noise with zero mean and varying standard deviations to simulate different noise levels.

The first scenario, referred to as \textit{board~volume}, imitates the popular calibration approach with a hand-held ChArUco board placed randomly in front of the cameras.
Simulated checkerboards are randomly distributed within a cylindrical working volume with a 3-meter radius and a maximum height of 1.5 meters.

In the second scenario, referred to as \textit{board~floor}, the checkerboards are randomly distributed on the floor, within a circular working area on the floor of a 3-meter radius, simulating a calibration board moved on the floor.
Both scenarios simulate a ChArUco board of size 120x85cm, which is identical to the one used in \cref{subsec:xp_calib_real} and roughly corresponds to the minimal size that is detectable by all cameras in practical conditions.
During the sampling of checkerboard poses, we ensure that the average number of observations per camera category is consistent across all scenarios, specifically around 2,000 for near-field and 3,000 for far-field cameras. 
The calibration is performed using the gold standard checkerboard method, which optimizes the board poses while constraining the points to align with the known board geometry.

In the third scenario, referred to as \textit{grid~floor}, 3200 3D points are distributed in a regular grid on the floor, within an area matching the one covered by the projector used in \cref{subsec:xp_calib_real}. This corresponds to the spatial distribution of the MSM marker centers. 
In this case, the cameras are calibrated using the algorithm described in \cref{subsec:alg}, without incorporating a coplanarity constraint, in order to apply to the most general acquisition scenario.

\noindent \textbf{Evaluation} The calibration quality is assessed by comparing the estimated camera poses with the ground truth. This is done by aligning the estimated point cloud to the ground truth using a \textit{Sim}(3) similarity that minimizes the squared error via Umeyama’s algorithm \citep{umeyama}.

\begin{figure}[t]
    \centering
 
    \begin{minipage}{0.49\textwidth}
        \centering
        \includegraphics[width=\textwidth]{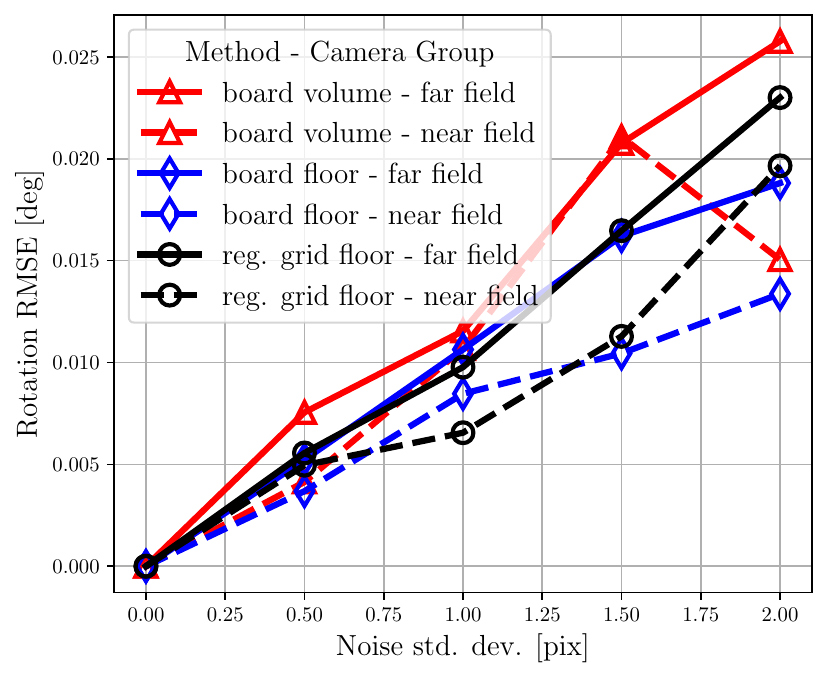}
        \subcaption{}
    \end{minipage}%
    \begin{minipage}{0.49\textwidth}
        \centering
        \includegraphics[width=\textwidth]{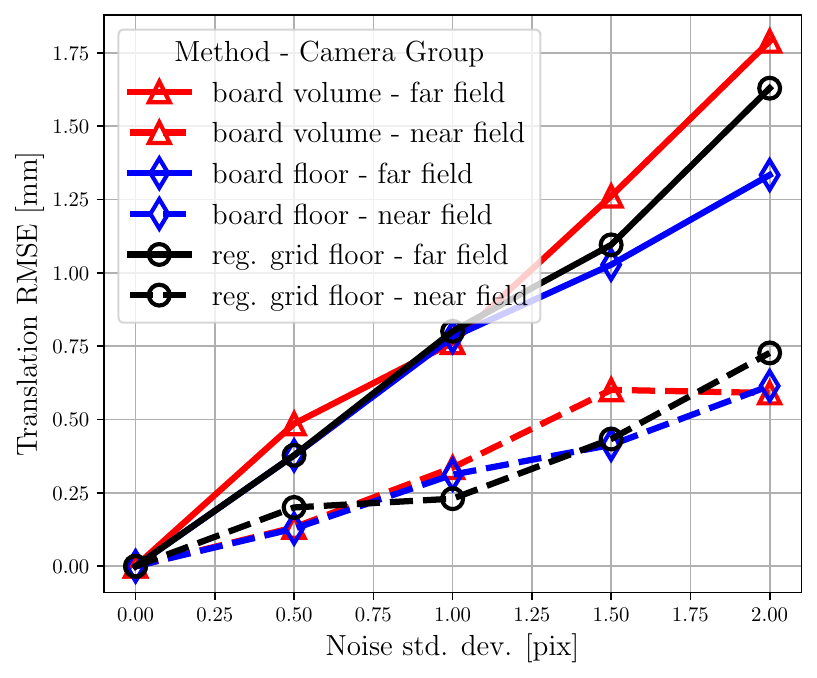}
        \subcaption{}
    \end{minipage}%
    
    \caption{(a) Rotation RMSE and (b) translation RMSE over 200 simulations. See text for details}
    \label{fig:eval_calib_synth_results}

\end{figure}

\noindent \textbf{Results} \cref{fig:eval_calib_synth_results} shows the rotation and translation RMSE over 200 simulations. For the same level of noise on the observations, the three scenarios deliver comparable results, indicating that a planar distribution of 3D points does not negatively impact the calibration quality. \revision{The higher error for the \textit{board volume} scenario can be attributed to its shorter average track length compared to the other scenarios. In this case, the random orientation of the board limits the visibility of the planar pattern to only a subset of cameras. This reduces the average track length, as fewer cameras can observe the same points. Given a fixed number of observations per camera, shorter track lengths increase the ratio of 3D points (unknowns) to constraints, resulting in less accurate estimates.} For the scenarios with coplanar points (\textit{board~floor}, \textit{grid~floor}), the results are also comparable. This indicates that, provided a sufficient number of points are used and uniformly distributed across the images, the additional constraints introduced by a board of known geometry in the \textit{board floor} scenario have a marginal impact on calibration accuracy.

\subsection{Calibration on real-world data}
\label{subsec:xp_calib_real}

We conducted real-world experiments in a mock-up OR to replicate the typical conditions of 3D-SSR. 

\noindent \textbf{Setup} The experimental setup (see \cref{fig:methodo_fig} (a)) included 9 cameras: 8 GoPro cameras (6 positioned in the far field and 2 mounted on surgical lamps near the surgical table in the near field) and 1 ceiling-mounted Canon camera with a strong optical zoom focusing on the surgical field. From this setup, we sampled different camera configurations to create four distinct setups with varying positions and viewpoints to test the robustness of the calibration method under setups of varying difficulty (\textit{Full}, \textit{Easy}, \textit{Medium}, \textit{Hard}). As we assume a (nearly) planar surface in our method, we removed the table during the calibration procedure. We compared our method to the gold standard multi-camera calibration method (same as used in \cref{subsec:xp_calib_synth}) using a moving ChArUco board. We selected the ChArUco board with the smallest square size that was detectable in all cameras, including those in the far field. The board was manually moved throughout the scene to gather sufficient inter-view correspondences. In order to not penalize this method, the table was also removed during the calibration procedure. The calibration process took approximately 5 minutes to complete. Additionally, we evaluated the state-of-the-art SfM pipelines COLMAP \citep{schonberger2016structure} and GLOMAP \citep{pan2024glomap} under two conditions: (i) in a fully furnished OR (including the table), and (ii) with a projected texture on the floor and the lights turned off to maximize contrast, with the OR table removed to avoid penalizing the method. For the proposed method, an entry-level projector (Acer H6518STi) was mounted on the OR ceiling. A sequence of 100 MSM arrays was projected onto the OR floor with the lights turned off. Each array contained 32 MSMs and  was slightly offset from the previous one to create a dense grid of 3200 points across the floor. Seven scale factors, $\lambda \in \Lambda = \{1.0, 1.4, 2.0, 3.0, 4.0, 6.0, 8.0\}$, were used in the experiment, corresponding to marker sizes ranging from 5 cm to 40 cm. The experiment was repeated for MSMs generated from two different marker systems: AprilTags and ArUco. The proposed calibration procedure in our experiment took 70 seconds. 

\noindent \textbf{Cameras}
The GoPro (Hero 12 Black) cameras have an approximate focal length of 915 pixels, while the ceiling-mounted Canon CR-N300 has a focal length of 11,100 pixels (strong optical zoom). All cameras recorded at a resolution of 1920x1080 (Full HD). The intrinsic parameters and distortion coefficients for all cameras were calibrated beforehand using a standard internal calibration procedure involving a checkerboard. For the proposed method, the recordings were made at 30 fps. For the ChArUco-based method, we employed a higher frame rate of 240 fps to minimize motion blur and temporal synchronization issues during the manual movement of the board, ensuring the method was not disadvantaged by these factors.

\noindent \textbf{Evaluation}
To ensure a meaningful evaluation, we recorded a separate dataset, independent from the calibration data, using the ChArUco board for generating accurate 2D-2D correspondences. This  dataset was used for evaluating the performance of all methods as follows. Given the pose estimates provided by a calibration method, we performed pairwise triangulation between cameras to obtain an initial estimate of the 3D point locations. Then, we performed a global non-linear refinement to minimize the total reprojection error across all cameras.

\begin{figure}[ht!]
    \centering
    \begin{minipage}{0.33\textwidth}
        \centering
        \includegraphics[width=0.98\textwidth]{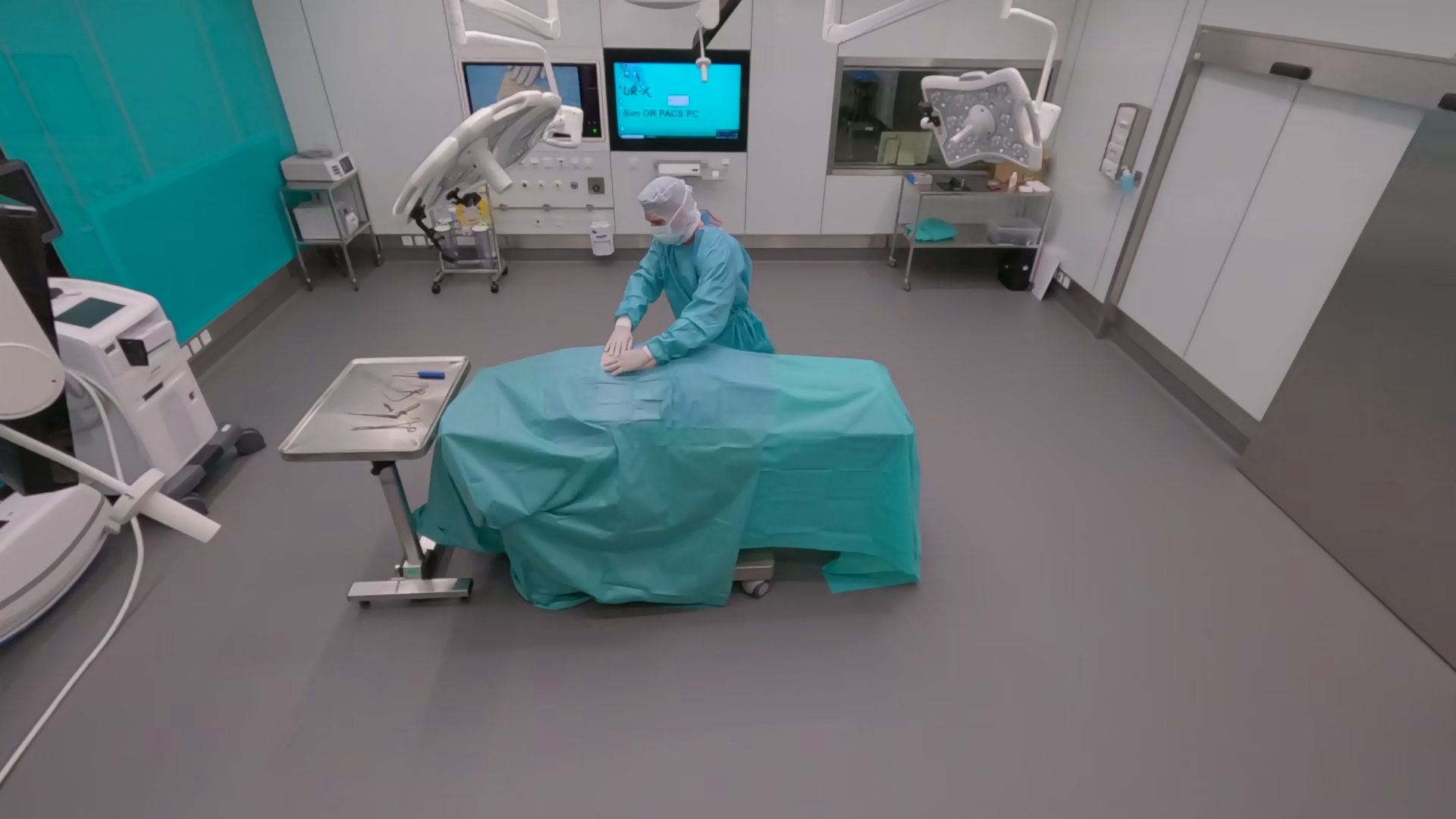}
    \end{minipage}%
    \begin{minipage}{0.33\textwidth}
        \centering
        \includegraphics[width=0.98\textwidth]{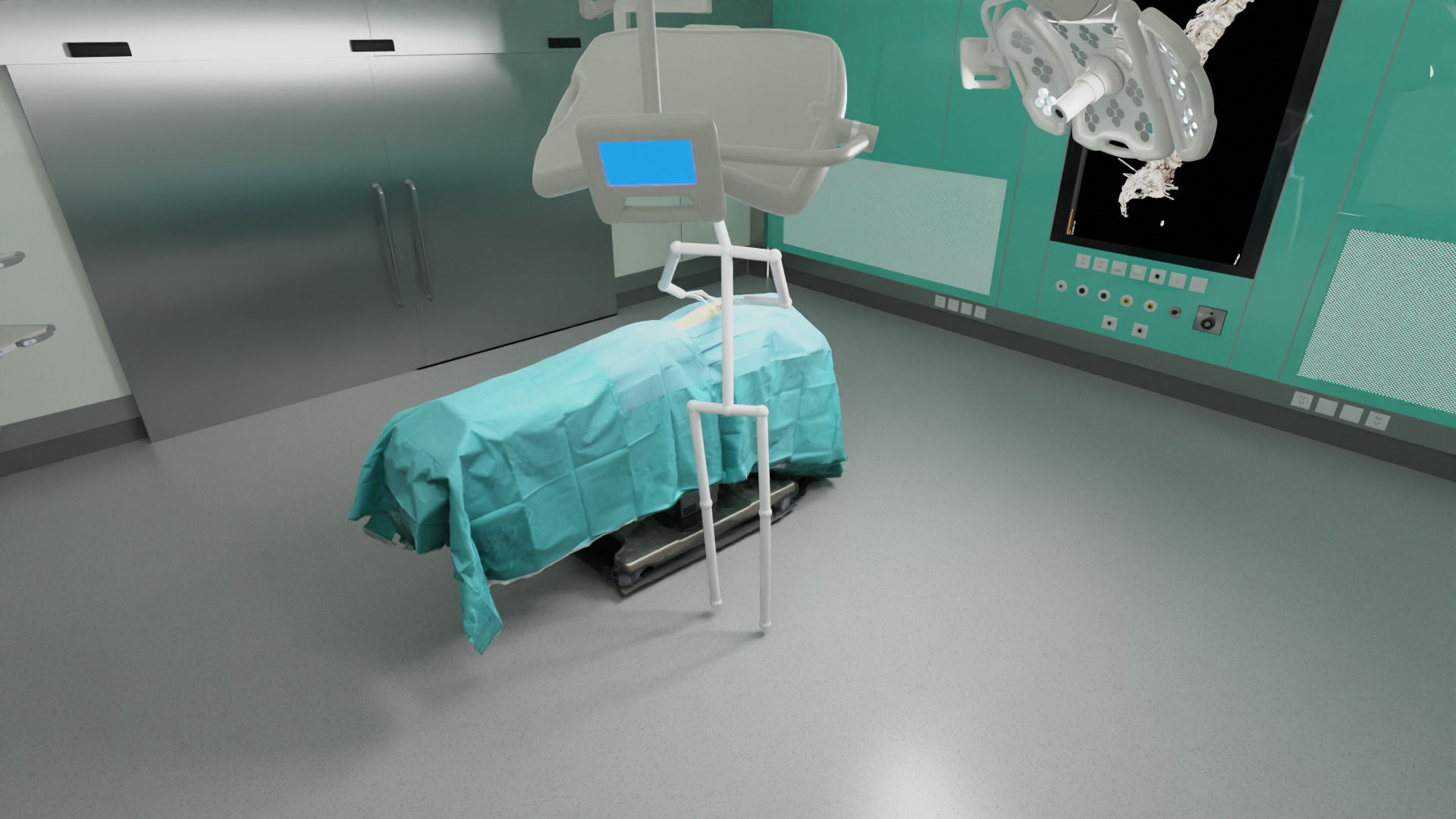}
    \end{minipage}%
    \begin{minipage}{0.33\textwidth}
        \centering
        \includegraphics[width=0.98\textwidth]{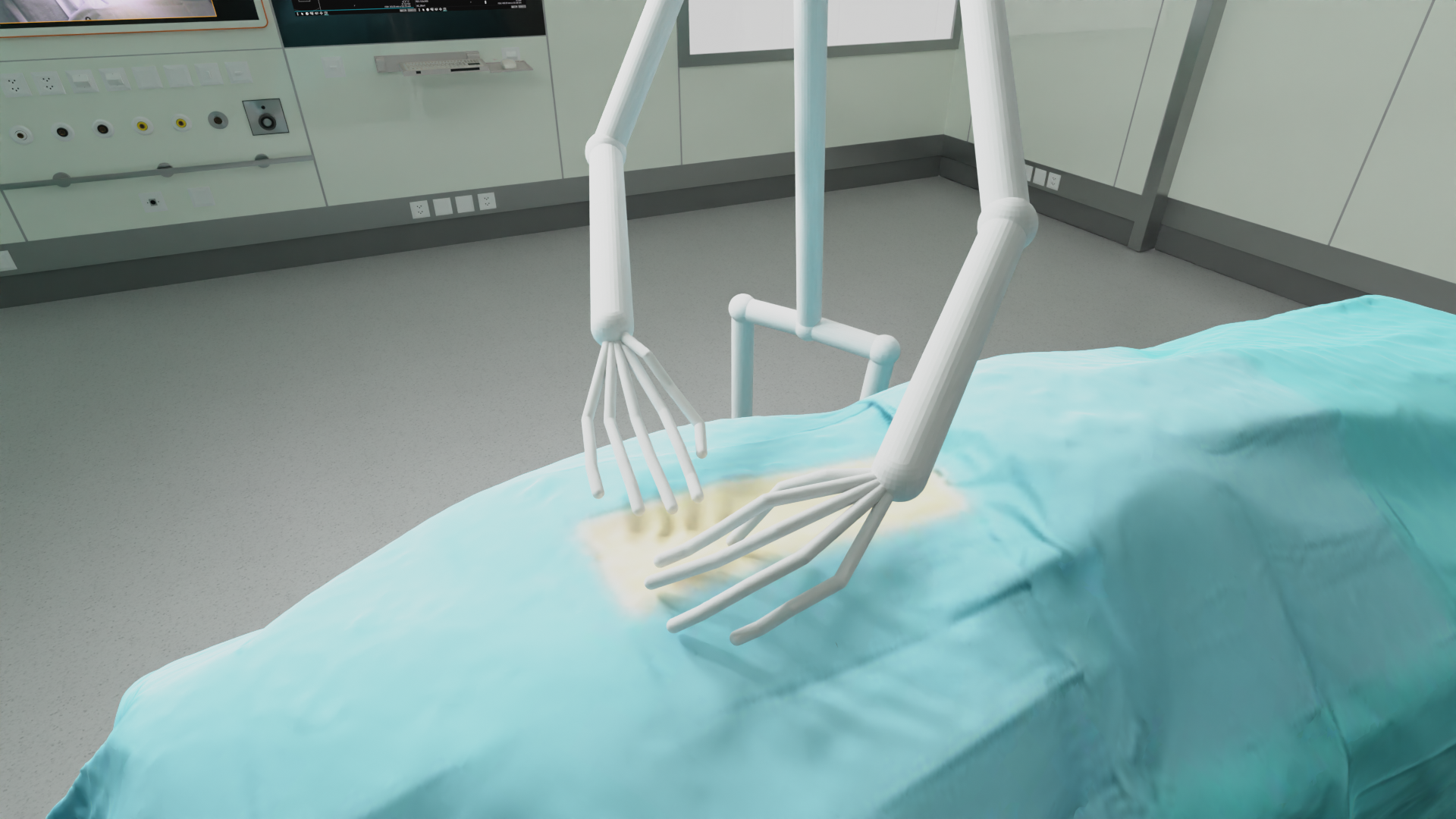}
    \end{minipage}%
    \caption{Application of our calibration method to 3D-SSR. From left to right: reference image, the surgeon's body poses, and hands, successfully reconstructed using camera poses computed with the proposed MSM-based calibration method. The reconstructed objects are integrated into the OR digital twin \citep{Hein_2024_CVPR}. An animation is available in Online Resource 2}
    \label{fig:applications}
\end{figure}

\noindent \textbf{Results} All results presented for our method use ArUco markers and integrate the coplanarity constraint in the optimization. Results without the coplanarity constraint are very comparable and available in Online Resource 1.
\cref{tab:res_all} presents the reprojection errors for each distinct setup, while more detailed results for each camera category for the \textit{full} setup are shown in \cref{tab:res_full}. \cref{tab:res_success_rates} evaluates the robustness by computing success rates for each camera setup, defined as the percentage of reconstructed cameras with a mean reprojection error below a threshold of 0.5/2/5 pixel on the evaluation dataset. \revision{We also compare the reconstructed point clouds and camera poses obtained using our method against those from the ChArUco baseline in the Full setup. The camera poses exhibit a rotational RMSE of 0.12 degrees and a translational RMSE of 6.13 mm, corresponding to a relative error of less than 0.14\% of the mean inter-camera distance (4.22 m). The mean Euclidean error on the 3D points in the evaluation dataset is 1.63 $\pm$ 0.91 mm.} 
Last, we show the accuracy of our calibration by applying it to 3D-SSR, as shown in \cref{fig:applications}. 
We successfully reconstruct the surgeon’s body and hands poses by locating the joints in the images and triangulating them using the cameras poses obtained with proposed calibration method. In that example, body joints were computed using \citep{openpose}, while hands joints were manually extracted.

\begin{table}[t]
\begin{tabular}{lcccccccc}
\hline 
 & \multicolumn{2}{c}{Full }        & \multicolumn{2}{c}{Easy }          & \multicolumn{2}{c}{Medium }        & \multicolumn{2}{c}{Hard} \\
 & ChAr. & Ours & ChAr. & Ours & ChAr. & Ours & ChAr. & Ours \\ \hline
Calibration & 0.33  & 0.35  & 0.30  & 0.34  & 0.35  & 0.32  & 0.30  & 0.23  \\ 
Evaluation  & 0.18  & 0.28  & 0.17  & 0.27  & 0.21  & 0.27  & 0.16  & 0.16  \\ \hline
\end{tabular}
\caption{Reprojection errors on calibration and evaluation data for the ChArUco-based and our proposed calibration method on distinct setups}

\label{tab:res_all}
\end{table}

\begin{table}[t]
\begin{tabular}{llcccccccc}
\hline
& & \multicolumn{2}{c}{Far Field} & \multicolumn{2}{c}{Near Field} & \multicolumn{2}{c}{Close-up} & \multicolumn{2}{c}{All} \\
& & \multicolumn{1}{l}{ChAr.} & Ours & \multicolumn{1}{l}{ChAr.} & Ours & \multicolumn{1}{l}{ChAr.} & Ours & \multicolumn{1}{l}{ChAr.} & Ours \\ \hline
\multicolumn{1}{c}{\multirow{3}{*}{Calib.}} & Repr. err.   & 0.32  & 0.35  & 0.40  & 0.30  & -  & 0.06  & 0.33  & 0.35 \\ 
\multicolumn{1}{c}{}                        & Distrib. score& 231   & 208   & 378   & 316   & -  & 376   & 268   & 251  \\ 
\multicolumn{1}{c}{}                        & \# obsv.     & 1172  & 3011  & 576   & 1301  & -  & 49    & 1023  & 2301 \\ 
\multicolumn{1}{l}{Eval.}                   & Repr. err.   & 0.19  & 0.30  & 0.12  & 0.15  & -  & 0.02  & 0.18  & 0.28 \\ \hline
\end{tabular}
\caption{Results per camera category on the \textit{full} setup for the ChArUco and our proposed method. Mean reprojection error (calib. \& eval.), disribution score (calibration data, see \cref{subsec:alg}), observations of 3D points with track length $>= 2$ (calibration data). 
The ChArUco-based method failed to register the close-up camera}
\label{tab:res_full}
\end{table}

\begin{table}[t]
\begin{tabular}{lcccccc}
\hline
       & \multicolumn{2}{c}{COLMAP}                  & \multicolumn{2}{c}{GLOMAP}                  & ChArUco     & Proposed    \\
       & \multicolumn{1}{c}{Lights on} & Proj. Text. & \multicolumn{1}{c}{Lights on} & Proj. Text. &             &             \\ \hline
Full   & \multicolumn{1}{c}{0/0/0}     & 0/33/89     & \multicolumn{1}{c}{0/0/22}    & 0/22/89     & 89/89/89    & 100/100/100 \\
Easy   & \multicolumn{1}{c}{0/0/0}     & 0/50/100    & \multicolumn{1}{c}{0/0/50}    & 0/0/83      & 100/100/100 & 100/100/100 \\
Medium & \multicolumn{1}{c}{0/0/0}     & 0/33/100    & \multicolumn{1}{c}{0/0/0}     & 0/33/83     & 100/100/100 & 100/100/100 \\
Hard   & \multicolumn{1}{c}{0/0/0}     & 0/0/0       & \multicolumn{1}{c}{0/0/0}     & 0/0/0       & 67/67/67    & 100/100/100 \\ \hline
\end{tabular}

\caption{Success rates in \% with a threshold of 0.5/2/5 pixel. See text for details
}
\label{tab:res_success_rates}

\end{table}

\section{Discussion}\label{sec:Discussion}
The proposed method achieves errors of the same order of magnitude as the manual ChArUco method, while demonstrating  better robustness under extreme viewpoint changes, as demonstrated in the \textit{Hard} configuration.
The difference between the ChArUco baseline and our method, approximately 0.1 pixel in reprojection error, may be attributed to thermal effects from prolonged camera usage.
The ChArUco method had a slight advantage because its calibration data were collected immediately before the evaluation dataset. In contrast, the data acquisition for the proposed method was performed about an hour earlier. 
Overall, the proposed method demonstrates higher robustness while being fully automated. It is also flexible and can be easily extended to incorporate multiple projectors. 
Regarding the required projection duration, our findings indicate that using $|\Lambda|=3$ MSM scales is sufficient in most scenarios, potentially reducing the calibration time to 30 seconds. 

\noindent\textbf{Limitations} While our calibration approach offers automation and acquisition speed, it still has some practical limitations, four of which are notable.  
First, our current method assumes a planar projection surface, which necessitates the removal of the operating table. 
This limitation is acceptable for calibrations performed before or between surgeries; however, it prevents intra-operative recalibration. 
The robustness of our method in scenarios where the operating table remains in place still needs to be evaluated. 
\revision{ Second, the MSMs projection in our experiment was conducted with the main OR lights turned off.
A professional-grade projector could provide sufficient brightness to operate under normal OR lighting conditions but is significantly larger than our entry-level projector.}
Third, it requires that each camera's field of view overlaps with the projection. We estimate that this limitation is negligible in most practical applications, as the patient and anatomy can only be captured from an elevated point-of-view in the vast majority of use-cases.
\revisisonSecondTour{Fourth, internal calibration must be performed beforehand.}

\section{Conclusion}\label{sec:Conclusion}
In this work, we presented a fully automated method for external calibration of multi-camera systems, specifically tailored to environments with challenging camera setups and limited overlapping fields of view, such as 3D-SSR. By utilizing a ceiling-mounted projector and multi-scale markers (MSMs), our approach achieves high accuracy, robustness, and speed, eliminating the need for manual intervention and pave the way for fully automated 3D-SSR. 
Future work will focus on extending the proposed method to a scenario including the OR table and increasing the acquisition speed by utilizing color channels to project multiple MSM arrays simultaneously. 
\section*{Acknowledgments}
This work has been supported by the OR-X - a swiss national research infrastructure for translational surgery - and by the University of Zurich and University Hospital Balgrist.
\section*{Declarations}

\noindent \textbf{Competing Interests} The authors have no relevant financial or non-financial interests to disclose.

\noindent \textbf{Ethics Approval}
Since this study did not involve human or animal subjects, ethics approval is not applicable.

\noindent \textbf{Informed Consent}
As this research did not involve human participants, informed consent is not applicable.

\bibliography{sn-bibliography}

\end{document}